\documentclass[12pt]{article}
\usepackage[utf8]{inputenc}
\usepackage[top=1in, bottom=1in, left=1in, right=1in]{geometry}
\usepackage[nodisplayskipstretch]{setspace}
\usepackage[authoryear,round]{natbib}
\usepackage{amsmath}
\usepackage{amssymb}
\usepackage{xcolor}
\usepackage{graphicx}
\usepackage{booktabs}
\usepackage{enumitem}
\usepackage{adjustbox}
\usepackage{float}
\usepackage{appendix}
\usepackage{graphicx}
\usepackage{calc} 
\usepackage{mathrsfs} 
\usepackage{adjustbox}
\usepackage{etoolbox}
\AtBeginEnvironment{tabular}{\small}

\newcommand\blfootnote[1]{
  \begingroup
  \renewcommand\thefootnote{}\footnote{#1}%
  \addtocounter{footnote}{-1}%
  \endgroup
}

\usepackage[compact]{titlesec}  
\titlespacing*\section{0pt}{4pt plus 4pt minus 2pt}{0pt plus 2pt minus 2pt}
\titlespacing*\subsection{0pt}{4pt plus 4pt minus 2pt}{1pt plus 2pt minus 2pt}

\newcommand{\indep}{\perp \!\!\! \perp}
\newcommand{\Ptest}{\mathbb{P}^*}
\newcommand{\PtestX}{\mathbb{P}_{X}^*}
\newcommand{\wtest}{w^*}
\newcommand{\wtesthat}{\hat{w}^*}
\newcommand{\thetatrue}{\theta^{\sharp}}
\newcommand{\pitrue}{\pi^{\sharp}}

\DeclareMathOperator{\E}{\mathbb{E}}
\DeclareMathOperator*{\esssup}{ess\,sup}
\newcommand*\mycdot{{\mkern 2mu\cdot\mkern 2mu}} 

\title{\vspace{-2em}Discussion of Kallus (2020) and Mo, Qi, and Liu (2020): New Objectives for Policy Learning}
\author{{\large Sijia Li$^{1,*}$, Xiudi Li$^{1,*}$, Alex Luedtke$^{2,\dagger}$} \\[0.75em]
$^1$Department of Biostatistics, University of Washington, Seattle, WA, USA\\
$^2$Department of Statistics, University of Washington, Seattle, WA, USA\\
$^*$These authors contributed equally to this work. Order alphabetical.\\ $^\dagger$To whom correspondence should be addressed. E-mail:  aluedtke@uw.edu.}
\date{}

\begin{document}

\maketitle
\blfootnote{This work was supported by the National Institutes of Health (NIH) under award number DP2-LM013340. The content is solely the responsibility of the authors and does not necessarily represent the official views of the NIH.}

\setstretch{1.45}

\vspace{-3em}
\section{Introduction}

We congratulate Kallus and Mo, Qi, and Liu for introducing thought-provoking new policy learning objective functions. 
Each of the proposed objectives involves an empirical estimate of the value function under one or more synthetic covariate shifts. These shifts are defined in such a way that the resulting policy estimator is expected to outperform standard practice in terms of a certain measure of performance. The two works differ in which performance measures they focus on and, consequently, in the form of the synthetic shift(s) considered.

To be able to give a unified discussion of their proposed approaches, we adopt the notation used in \cite{kallus2020more} throughout. In particular, $(X,A,Y)$ is a triplet containing the covariate, action, and outcome, $\mathbb{P}$ denotes the distribution that gave rise to the independent and identically distributed (iid) training dataset, $\mathbb{E}$ denotes expectations under this distribution, $\pi(\mycdot|x)$ is a stochastic policy that gives a distribution over actions for a given covariate value, $\mu(a,x):=\E[Y|A=a,X=x]$ is the outcome regression, $\phi(a|x):= \mathbb{P}(A=a|X=x)$ is the propensity score, and $\sigma^2(a,x) := \text{Var}(Y|A=a,X=x)$ is the conditional variance of the outcome. For any distribution $\mathbb{Q}$ of $(X,A,Y)$, we let $\mathbb{Q}_X$ denote the marginal distribution of $X$ under sampling from $\mathbb{Q}$.





Most of Kallus' piece does not explicitly focus on the problem of covariate shift between a training and testing set. Instead, its main focus
is to find a policy learning objective that is easier to estimate than one that is commonly used in practice, namely the value function $V(\pi):=\E[\sum_a \pi(a|X) \mu(a,X)]$. To find such an objective, Kallus leverages the observation that, provided the policy class $\Pi_0$ contains an unconstrained optimal policy, the set of maximizers of any positively-weighted value function $V(\pi;w):= \E[\sum_a w(X)\pi(a|X) \mu(a,X)]$ over $\pi\in\Pi_0$ does not depend on the choice of weight $w$. Kallus then derives a weight function $w$ that minimizes a term that appears in the variance of an efficient estimator of (a centered version of) the weighted value function, subject to the constraint that $\E[w(X)]=1$. 
Under this constraint, this objective function can also be interpreted as the value of the rule $\pi$ in the population with the same outcome regressions $\mu$ as were observed under $\mathbb{P}$, but in which the density of the covariates is shifted from $p(x)$ to $q(x):= w(x)p(x)$.

Mo et al. consider the covariate shift problem more directly. These authors specify a collection of possible testing covariate distributions $\mathcal{Q}_c$ and measure the performance of each policy according to the least favorable shift in this collection. 
More concretely, they aim to learn a maximizer of $\inf_{w\in\mathcal{W}_c}V(\pi;w)$ over $\pi\in\Pi_0$, where $\mathcal{W}_c:=\{d\mathbb{Q}_X/d\mathbb{P}_X : \mathbb{Q}_X\in\mathcal{Q}_c\}$. The sets $\mathcal{Q}_c$ are defined in such a way that $\mathcal{Q}_c$ is a singleton that contains the training covariate distribution $\mathbb{P}_X$ when the distributional robustness constant $c$ is equal to 1, and $\mathcal{Q}_c$ grows as $c$ grows.  Selecting the tuning parameter $c$ enables a tradeoff between learning a policy that will perform well at the training distribution and learning a policy that will perform well even if there is a covariate shift in the testing distribution.

In Section~\ref{sec:Kallus}, we discuss Kallus' elegant retargeted policy learning framework. We make the following observations:
\begin{enumerate}[label=Sec~2.\arabic*:,labelwidth=\widthof{\ref{last-item}},leftmargin=!,noitemsep,topsep=0em]
    \item We show that a variety of constraints are possible in the optimization problem considered by Kallus. Moreover, we show that the constraint imposed in the piece under discussion, namely that $\E[w(X)]=1$, may have a meaningful impact on the chosen weight. This impact does not appear to have been explored in that work.
    \item We show that, under appropriate regularity conditions, it is necessary to consider the curvature of the value function to be able to relate the regret in a retargeted population to the regret in a testing population. This suggests that the weight presented in Kallus, which does not take this curvature into account, may perform poorly relative to standard practice in certain settings.
    \item We introduce two alternative constraints that take this curvature into account.
    \item\label{last-item} We use a simulation study to show that ignoring curvature when selecting a weight function can lead to poor performance. This study also demonstrates the promise of methods that take this curvature into account.
\end{enumerate}

\noindent In Section~\ref{sec:Mo}, we discuss Mo et al.'s inspiring proposal. 
We make the following observations: 
\begin{enumerate}[label=Sec~3.\arabic*:,labelwidth=\widthof{\ref{last-item2}},leftmargin=!,noitemsep,topsep=0em]
    \item The authors proposed using a (possibly small) calibration sample from the testing distribution of interest to determine the size of the uncertainty set around the training distribution. We discuss the alternative possibility of centering this set around this testing distribution, which we suspect 
    will be less conservative under large covariate shift.
    \item\label{it:MoEstimators} The proposed estimators for the value function 
    might suffer from suboptimal precision. We propose alternative estimators in settings where a calibration sample is available, including the estimators that are efficient within the covariate-shift statistical model considered by Mo et al. 
    \item\label{last-item2} We show via simulation that using our proposed estimators can lead to dramatic improvements in precision for a range of calibration sample sizes and regardless of whether there is, in fact, covariate shift in the calibrating distribution.
\end{enumerate}

\section{Discussion of \cite{kallus2020more}} \label{sec:Kallus}

\subsection{Impact of the chosen scaling constraint on the optimal weight}\label{sec:constraint}

Kallus proposes to choose an optimal (positive) weight function $w_0$ and optimal reference policy $\rho_0$ to solve the following optimization problem:
\begin{align}
    &\textnormal{minimize }&&\Omega(w,\rho) := \mathbb{E}\left[w^2(X)\left(\sum_{a \in \mathcal{A}}\tfrac{\sigma^2(a,X)}{\phi(a|X)}\rho^2(a|X)+\max_{a\in\mathcal{A}}\tfrac{\sigma^2(a,X)}{\phi(a|X)}\{1-2\rho(a|X)\}\right)\right] \nonumber \\
    &\textnormal{subject to }&&\mathbb{E}[w(X)] = 1.\label{eq:constraint}
\end{align}
We will refer to this optimal weight as ``the retargeting weight". Let $\mathbb{P}_{X}$ denote the covariate distribution in the training population. 
Because $\Omega(\mycdot,\rho)$ is not scale-invariant, it appears to be important to impose some form of constraint; without any constraint, there is no positive weight function that solves the above --- indeed, the objective function can always be made smaller by simply rescaling the weight function to be closer to zero.

\begin{figure}
    \centering
    \includegraphics[width=0.9\textwidth]{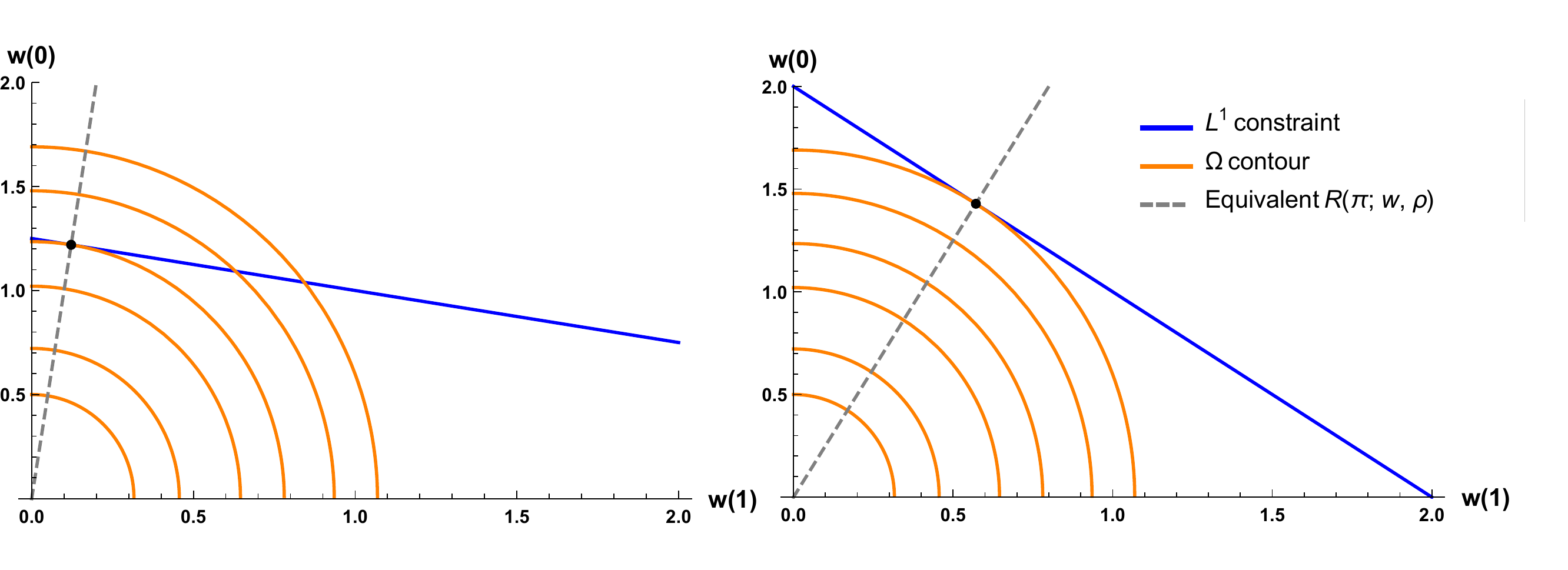}
    \caption{Optimal weight in the simple example with different measures $\nu$ in the $L^1(\nu)$ norm constraints. Left panel: $\nu=\mathbb{P}_X = $ Bernoulli(0.2), optimal weight $\approx (1.22, 0.12)$; Right panel: $\nu = $ Unif$\{0,1\}$, optimal weight $\approx (1.43, 0.57)$. By the homogeneity of $w\mapsto R(\pi;w,\rho)$, weights along the dashed lines yield equivalent policy learning problems.} 
    \label{contour}
\end{figure}

There are a variety of constraints that can be imposed to avoid this challenge. For example, this problem is still avoided if the constraint in \eqref{eq:constraint} is replaced by any constraint of the form $\|w\|_{L^1(\nu)}=1$, where $\nu$ is some user-specified measure. In fact, the constraint in \eqref{eq:constraint} is a special case of such a constraint when $\nu=\mathbb{P}_X$. 
Given the wide variety of possible constraints, it is natural to wonder whether there is any reason to prefer any given one of them. Our first intuition was that, because the constraint is only being employed to set the scale for the objective function, the resulting policy learning problem would not depend on the particular constraint that is imposed. However, after further consideration, we realized that our intuition was incorrect --- in fact, different choices of the dominating measure can lead to meaningfully different optima, which can in turn lead to meaningfully different policy estimators. To illustrate this, we consider a simple case where the covariate distribution $\mathbb{P}_X$ in the training population is Bernoulli$(0.2)$ and the action is binary. We take $\phi \equiv 0.5$, $\sigma^2(a,0) = 0.1$ and $\sigma^2(a,1) = 1$ for $a \in \{-1, 1\}$. The contours of the objective take the form of ellipses. In Figure \ref{contour}, we take the dominating measure $\nu$ to be either the training distribution $\mathbb{P}_X$ or a uniform distribution over $\{0,1\}$, and we observe that they lead to different optimal weights. Specifically, under $\mathbb{P}_X$, the vast majority of the weight is placed at $X=0$, whereas the weight is more evenly distributed across $X=0$ and $X=1$ under the uniform distribution. It is not immediately clear whether there is a reason to prefer one of these two weight functions over the other. 


One argument that we could imagine in favor of the constraint employed in the work under discussion, namely the $L^1(\mathbb{P}_X)$ constraint, is that the optimal weight $w_0$ is a density ratio between a hypothetical population with density $w_0(x)p(x)$ and the training population with density $p(x)$. However, as we will now show, any $\mathbb{P}_X$-integrable weight function can be rescaled to have this interpretation. For a functional $f$ of a positive weight function $w$, we will call $f$ (strictly positively) homogeneous if $f(kw)=kf(w)$ for all $k>0$ and all $w$. Now, by the homogeneity of $w\mapsto R(\pi;w,\rho):=V(\pi;w)-V(\rho;w)$, for any $\mathbb{P}_X$-integrable $w$, the weight function $\tilde{w}(x):= w(x)/\E[w(X)]$ will lead to an equivalent policy learning problem to that which arises under $w$. 
Moreover, like the retargeting weight introduced by Kallus, $\tilde{w}$ is a density ratio. Therefore, the question remains as to whether there is a compelling reason to prefer the formulation of the constraint in \eqref{eq:constraint} to other possibilities. We will return to this question in Section~\ref{sec:curvature}.




\subsection{Relating the weighted regret to the testing regret}

For a given policy $\pi$ and weight $w$, let $\mathcal{R}(\pi;w):=\sup_{\pitrue\in\Pi_0} V(\pitrue;w) - V(\pi;w)$ denote the regret of deploying $\pi$ in the $w$-weighted population as opposed to deploying an optimal policy. Here we consider the case that the policy will be deployed in the testing population that has the same distribution of $Y|A,X$ as does $\mathbb{P}$ and whose covariate distribution is shifted to $\mathbb{P}_X^*$, which is mutually absolutely continuous with respect to $\mathbb{P}_X$. The value of deploying a policy $\pi$ in this population is equal to $V(\pi;w^*)$, where the testing weight $w^*$ is defined to be equal to $d\mathbb{P}_X^*/d\mathbb{P}_X$. 
As is shown in Eq.~8 of the work under discussion, for any $w$, a bound is available on the regret $\mathcal{R}(\hat{\pi}^w;w)$ of the policy $\hat{\pi}^w$ learned using weight $w$ that scales with the square root of $\Omega(w,\rho_0)$. 
Following that display, the author relates this bound to the regret in the testing population 
via the following expression, which holds for any policy $\pi$: 
\begin{align}
    \mathcal{R}(\pi;w^*)&\le \left[\esssup_x w^*(x)/w(x)\right] \mathcal{R}(\pi;w). \label{eq:KallusRwbd}
\end{align}
The author states that this bound is too loose to accurately characterize the regret on the original population ($w^*=\boldsymbol{1}$). We will now argue that, on the contrary, this bound may actually be highly informative in some settings, regardless of whether or not the testing population corresponds to the original population. We will then show that this suggests that the policy that is learned based on the retargeting weight may perform quite poorly when it places much less weight on certain covariate values than does the testing weight.

To show this, we will work in a simple setting where the well-specified class $\Pi_0$ is indexed by a one-dimensional parameter $\theta$ --- that is, $\Pi_0 = \{\pi_\theta:\theta\in\Theta\}$ --- and there is a unique policy $\pi_{\thetatrue}$ that maximizes $V(\mycdot;w^*)$. Though much simpler than the general settings considered in many theoretical works on policy learning, focusing on this simple special case makes it possible to get a tight understanding of the relationship between the regrets in the testing and the retargeted populations. Moreover, it seems reasonable to ask that a new method should perform competitively with the common practice of using uniform weighting in such simple settings.

Lemma~2.1 in \cite{kallus2020more} shows that, when $\Pi_0$ is correctly specified, $\thetatrue$ is optimal for any choice of weight $w$. Thus, if $\theta\mapsto V(\pi_\theta;w)$ is twice continuously differentiable at $\thetatrue$, a Taylor expansion of $V(\mycdot;w)$ and the fact that $V'(\thetatrue;w)=0$ yield that
\begin{equation*}
    V(\pi_{\theta};w) - V(\pi_{\thetatrue};w) = \frac{V''(\thetatrue;w)}{2}(\theta-\thetatrue)^2 + o([\theta-\thetatrue]^2),
\end{equation*}
where we use little-oh terms to denote behavior as $\theta\rightarrow\thetatrue$. Applying the above at an arbitrary weight function $w$ and at the testing weight shows it is reasonable to expect that
\begin{equation}\label{regrets}
   \mathcal{R}(\pi_{\theta};w^*) = \left[\frac{V''(\thetatrue;w^*)}{V''(\thetatrue;w)}+o(1)\right]\mathcal{R}(\pi_{\theta};w)
\end{equation}
under appropriate conditions. The second derivative in the denominator above measures the curvature of the value surface as a function of $\theta$ when we use the weight $w$. 
If $\theta$ were multi-dimensional, a similar display to the above could be derived, but it would take the form of an upper bound involving the eigenvalues of the Hessians of $V(\mycdot;w^*)$ and $V(\mycdot;w)$.

We now use \eqref{regrets} to argue that, under certain conditions, for any given $\phi$ and $\sigma^2$, 
it is possible to construct a distribution with propensity $\phi$ and conditional variance function $\sigma^2$ such that \eqref{eq:KallusRwbd} is tight in first order for the resulting retargeting weight $w_0$. 
When doing this, we focus on the special case where the action is binary, the covariate is real-valued, and the simple policy class of the form $\pi_\theta(1|x) = I\{x > \theta\}$ is well-specified. Fix $\phi$ and $\sigma^2$, and note that, as the retargeting weight $w_0$ only depends on these two quantities, this weight is also fixed. Let $C(x):= \mu(1,x)-\mu(-1,x)$. 
Under smoothness conditions, $C(\thetatrue)=0$, and also $V''(\thetatrue;w)
     = w(\thetatrue)p(\thetatrue)[-C'(\thetatrue)]$. 
In this case, Eq.~\eqref{regrets} rewrites as $\mathcal{R}(\pi_\theta;w^*) = [w^*(\thetatrue)/w(\thetatrue)+o(1)]\mathcal{R}(\pi_\theta;w)$. 
Based on this expression, we now argue that \eqref{eq:KallusRwbd} can be quite tight for all $\theta$ in a neighborhood of $\thetatrue$. We start by noting that \eqref{eq:KallusRwbd} is tight in first order when the value of $w^*(\thetatrue)/w(\thetatrue)$ is close to the essential supremum of the weight ratio $w^*/w$. 
In models where the outcome regression is variationally independent \citep{van2003unified} of the propensity and conditional outcome variance, we can choose $\mu$, which entirely determines $\thetatrue$, such that $w^*(\thetatrue)/w(\thetatrue)$ is large. 
Though such variational independence does not always hold --- for example, it fails when the outcome is binary since $\sigma^2(a,x)=\mu(a,x)[1-\mu(a,x)]$ --- it will hold in many nonparametric models, such as most of those in which the outcome is continuous. Therefore, it is often reasonable to expect that, for any given instance of the retargeting weight, \eqref{eq:KallusRwbd} will be tight in first order when the outcome regression $\mu$ is unfavorable. 

Consequently, when the retargeting weight deviates dramatically from the testing weight, there will often exist outcome regressions that lead to the retargeted value surface being flat near the optimal policy as compared to the testing value surface; this, in turn, implies that, for any given retargeting weight $w_0$, there will often be cases in which there is a tenuous relationship between the retargeted regret $\mathcal{R}(\pi_\theta;w_0)$, which corresponds to the quantity that the retargeting estimator seeks to optimize, and the regret $\mathcal{R}(\pi_\theta;w^*)$ that is of substantive interest. Moreover, in cases where $w^*$ is known --- for example, because the testing population is known to coincide with the training population ($\mathbb{P}^*=\mathbb{P}$) --- the simple alternative approach that directly maximizes an estimate of the $w^*$-weighted value function appears to avoid this worst-case curvature issue altogether. This issue appears to be similarly avoidable when $w^*$ is not known but a calibration sample from $\mathbb{P}_X^*$ is available to use to estimate this quantity.

\subsection{Retargeting using curvature-based constraints}\label{sec:curvature}

The arguments at the end of the preceding subsection appear to suggest that, for unfavorable outcome regressions $\mu$, using a weight other than $w^*$ may have the undesirable consequence of flattening the value surface. 
However, in practice, information about the outcome regressions contained in the training data may all but rule out these unfavorable cases. Therefore, 
we now consider the setting in which the data-generating distribution, which may depend on sample size, is not chosen adversarially so as to flatten the value surface that results from choosing to use a particular weight. Even under this non-adversarial setup, \eqref{regrets} shows that curvature plays an important role in relating the regret in the retargeted population to that in the testing population. 
Motivated by this, we propose two curvature-based constraints that provide alternatives to the $L^1(\mathbb{P}_X)$ constraint in \eqref{eq:constraint}.

For the first of these constraints, we focus on the same simple case as we did in the previous section, namely that where $\Pi_0$ is indexed by a one-dimensional parameter $\theta$. To motivate this constraint, note that Kallus' Eq.~8 implies that, for a fixed class $\Pi_0$, 
\begin{align*}
    \mathcal{R}(\hat{\pi}^w;w)&= O_p\Big(\Omega(w,\rho_0)^{1/2}/n^{1/2} + \E[w(X)^2M(X)^2]^{1/2}/n^{1/2}\Big),
\end{align*}
where $M(x):= \max_a \mu(a,x)-\min_a \mu(a,x)$. Kallus argued the latter term in the big-oh expression above can be neglected for the purpose of selecting an optimal weight. Provided this latter term can be neglected, then, combined with \eqref{regrets}, this shows that we should expect that $\mathcal{R}(\hat{\pi}^w;w^*)$ will be smallest when $V''(\thetatrue;w^*)\Omega(w,\rho_0)^{1/2}/V''(\thetatrue;w)$ is smallest. It is worth noting that the accuracy of this bound is not a certainty, as \eqref{regrets} is only expected to be applicable if the parameter $\hat{\theta}^w$ indexing $\hat{\pi}^w$ is close to $\thetatrue$, and $\hat{\theta}^w$ is not expected to be close to $\thetatrue$ at local alternatives where $M$ converges to zero too quickly with sample size --- given works in other settings \citep[e.g.,][]{hirano2009asymptotics}, it seems likely that this fails to hold when $\E[w(X)^2M(X)^2]=O(1/n)$. Therefore, it seems like it could only be plausible that both $\E[w(X)^2M(X)^2]$ is negligible and $\hat{\theta}^w$ is close to $\thetatrue$, so that \eqref{regrets} is applicable, if it is true that, as $n\rightarrow\infty$, $\E[w(X)^2M(X)^2]\rightarrow 0$ while $n\E[w(X)^2M(X)^2]\rightarrow\infty$.

In any case, here we explore the possibility of choosing $(w^\dagger,\rho^\dagger)$ to minimize this heuristic bound $V''(\thetatrue;w^*)\Omega(w,\rho_0)^{1/2}/V''(\thetatrue;w)$ on $\mathcal{R}(\hat{\pi}^w;w^*)$. Using the monotonicity of the square root function, the concavity of $V(\mycdot;w^*)$ and $V(\mycdot;w)$ at $\thetatrue$, and the homogeneity of $w\mapsto V''(\thetatrue;w)$, we see that this is equivalent to choosing $(w^\dagger,\rho^\dagger)$ as a solution to
\begin{align}
    &\textnormal{minimize }\hspace{.5em}\Omega(w,\rho)\hspace{1em}\textnormal{subject to }\hspace{.5em}V''(\thetatrue;w) = -1.\label{eq:newconstraint}
\end{align}
While the reference policy $\rho^\dagger$ is the same as the reference policy $\rho_0$ from Lemma 3.4 in the work under discussion, the weight $w^\dagger$ will generally differ from the retargeting weight $w_0$.  

The form of the constraint in \eqref{eq:newconstraint} depends on the choice of $\Pi_0$. For example, as shown earlier when the action is binary and $\pi(1|x) = I\{x > \theta\}$, $V''(\thetatrue,w) \propto -w(\thetatrue)$. 
However, since the new constraint depends on $w$ locally, the optimal $w$ will put all of the mass at one point, namely $\thetatrue$. To overcome this difficulty, we propose to optimize $w$ over a restricted class of weights. 
A simple parametric form for the weight is given by $w_t(x) \propto (1-t)w_0(x) + t$ for $t\in [0,1]$, where $w_0(x)$ is the proposed retargeting weight and the normalizing constant is chosen so that the constraint in \eqref{eq:newconstraint} is satisfied. This form represents an interpolation between the retargeting weight and the uniform weight $\boldsymbol{1}$, which is commonly used in practice. We expect that when one choice significantly outperforms the other, we will pick the better-performing constraint by setting $t=0$ or $t=1$. 
More flexible parametric functional forms for $w$ 
could also be considered.

For the second constraint, we aim to find a form of $L^1$ constraint, along the lines of what we discussed in Section~\ref{sec:constraint}, that resembles a curvature constraint as well. Instead of the second derivative in \eqref{eq:newconstraint}, which measures curvature locally, here we consider a more global measure. Specifically, for the unconstrained policy class $\Pi$, we consider the maximum difference between values of two arbitrary policies when the population is weighted by $w$:
\begin{equation*}
    \sup_{\pi \in \Pi}V(\pi;w) - \inf_{\pi\in\Pi}V(\pi;w) = \E[w(X)M(X)].
\end{equation*}
A similar quantity appears in Kallus' Lemma 5.4 to characterize the probability that an optimal policy is chosen in the case where $\Pi_0$ is of finite cardinality. Kallus explains that this quantity measures ``the separation that a particular $w$ induces on the empirical value of different policies". Here we build on this idea and use a population-level analogue as our constraint. Let $\nu_1$ be a covariate distribution with density proportional to $M(x)p(x)$. We choose $(w^{\ddagger},\rho^{\ddagger})$ as the solution to
\begin{equation}\label{eq:secondconstraint}
     \textnormal{minimize }\hspace{.5em}\Omega(w,\rho)\hspace{1em}\textnormal{subject to }\hspace{.5em}\E_{\nu_1}[w(X)] = 1.
\end{equation}
The weight $w^{\ddagger}$ leads to the most easily estimable value function in terms of variance, among all the weights that give the same global measure of curvature. Compared to the first constraint, \eqref{eq:secondconstraint} has the advantage that it neither requires a working form for the weight function nor a constrained policy class $\Pi_0$ indexed by finite-dimensional parameters.
 
Like for the optimization \eqref{eq:constraint}, the solutions to \eqref{eq:secondconstraint} are available in closed form. In particular, similar arguments to those used to prove Kallus' Lemma 3.3 and Lemma 3.4 show that, 
when $|\mathcal{A}|=m$ and letting $\xi(x) := (m-2)\left[\sum_{a\in\mathcal{A}}\phi(a|x)/\sigma^2(a,x)\right]^{-1}$, they take the form
\begin{align}
    \rho^{\ddagger}(a|x) &= \frac{1}{2}\left(1- \frac{\phi(a|x)}{\sigma^2(a,x)}\xi(x)\right), \hspace{2em}  w^{\ddagger}(x) \propto M(x)\left(\sum_{a\in\mathcal{A}}\frac{\sigma^2(a,x)}{\phi(a|x)}+\frac{\xi(x)}{2}\right)^{-1}. \label{eq:thirdconstraint}
\end{align}
At the end of this subsection, we will compare this weight to the weight $w^\dagger$ that is optimal under the local curvature constraint.

To estimate the optimal weight $w^{\dagger}$ or $w^{\ddagger}$ from \eqref{eq:newconstraint} or \eqref{eq:secondconstraint}, we need to replace the unknown quantities by their estimates. In both cases, estimates of $\sigma^2$, $\phi$, and $\mu$ are required, either to learn the weight or to evaluate the resulting retargeted policy estimator. As \eqref{eq:secondconstraint} only involves these nuisance functions, there is no need to estimate additional quantities to learn $w^\ddagger$ or evaluate the resulting policy estimator. In contrast, \eqref{eq:newconstraint} depends on the unknown $\thetatrue$ as well. A simple solution is to replace it with some initial estimate coming from, for example, the uniform weight or the retargeting weight $w_0$. However, we observed in a simulation study that the resulting estimate $\hat w^\dagger$ is quite sensitive to the particular weight used to obtain this initial estimate. Specifically, we tried two initial estimates for $\thetatrue$ --- one was based on the retargeting weight, and the other was based on the uniform weight. When the retargeting weight was used to select the initial $\thetatrue$, the interpolation constant $t$ was typically chosen to be 0, so that the final selected weight $\hat w^{\dagger}$ corresponded to the retargeting weight. When the uniform weight was used, $t=1$ was typically selected, so that $\hat w^{\dagger}$ generally coincided with the uniform weight. Thus, the $t$ used in the initialization all but determined the final choice of $t$. 
How to deal with the unknown $\thetatrue$ is an interesting question for future work. In the case of the interpolation weight, or any other working forms indexed by some parameter $t$, one possibility is to choose $t$ to minimize $\hat\Omega(w_t,\hat\rho^{\dagger})$ subject to $\hat{V}''(\hat\theta_{w_t};w_t) =-1$, where $\hat\Omega$, $\hat V''$ are estimates of $\Omega$, $V''$ and $\hat\theta_{w_t}$ is an initial estimate of $\theta$ using weight $w_t$. Another, more general, approach is to replace the second derivative in \eqref{eq:newconstraint} with its infimum over a confidence interval for $\thetatrue$, when such an interval is available.

We now compare the behavior of the curvature-based weights $w^\dagger$ and $w^\ddagger$ in the special case where the policy class takes the form $\pi_\theta(1|x)=I\{x>\theta\}$. The expression for $w^{\ddagger}$ in \eqref{eq:thirdconstraint} suggests that larger weight should be placed where the difference among expected outcomes under different actions is larger, that is, where $M$ is large. This has the effect of making it easy to distinguish between the very best and worst policies in $\Pi$ and, consequently, making it unlikely that the learned policy will be substantially suboptimal. 
In contrast, $w^{\dagger}$ puts more weight near $\thetatrue$, where $M$ is in fact small. As noted above, in practice, $\thetatrue$ is unknown, and so an initial estimator must be used. On the one hand, if that estimator accurately locates the region in which $\thetatrue$ lies, then, under smoothness conditions, focusing weight there makes it possible to refine and improve the performance of the initial estimator. On the other hand, if the initial estimator is poor, then $w^\dagger$ may misdirect attention to a region that is not informative about the optimal policy learning problem, thereby resulting in a poor policy estimator. In future research, it would be interesting to study whether one of these divergent strategies should generally be preferred over the other.

\subsection{Simulation study}\label{subsec:kallussim}


We consider a binary action with $\mathcal{A}=\{-1,1\}$, focus on the case that the testing population coincides with the original population ($w^*=\boldsymbol{1}$), and generate data as follows: $X \sim \text{Unif}[-1,1]$, 
$P(A=1|X) = \phi(X)$,
and $Y|A,X \sim N[f_A(X),\sigma^2(X)]$, where $f_{-1}(x) = x$, and $f_1$, $\sigma^2$, and $\phi$ differ across the four scenarios that we consider. Specifically, we consider the following scenarios: (1) $f_1(x) = x + (x-0.5)I\{x > 0\}$; $\sigma^2(x) = 1$; $\phi(x) = 0.5I\{x \leq 0\}+\Phi(3.5x)I\{x>0\},$ where $\Phi$ is the cdf of a standard normal distribution; (2) $f_1(x) = x + (x-0.5)I\{x > 0\}$; $\sigma^2(x) = 0.01 I\{x \leq 0\}+I\{x>0\}$; $\phi(x) = 0.5$; (3) $f_1(x) = 2x$; $\sigma^2(x) = 1$; $\phi(x) = \Phi(3.5x)$; and (4) $f_1(x) = x+(x-0.3)I\{x>-0.4\}$; $\sigma^2(x) = 1$; $\phi(x) = 0.5I\{x \leq -0.4\}+\Phi(2.5x)I\{x>-0.4\}$. In scenarios 1 and 2, the conditional average treatment effect is 0 when $x \leq 0$, and hence only observations with positive covariate values are informative about the form of the optimal policy. However, when the covariates are positive, there is either poor overlap (scenario 1) or large outcome variance (scenario 2), and therefore these covariate values receive little weight under the retargeting scheme. This results in a flatter retargeted value surface around $\thetatrue$. In contrast, in scenario 3, which is similar to Kallus' simulation setting when his $\beta$ parameter equals to $3.5$, the outcome errors are homoskedastic and $\thetatrue$ lies in a neighborhood with good overlap, and so retargeting increases the curvature of the value surface. Finally, in scenario 4, the curvature under the retargeting weight and uniform weight is similar.


For the constraint in \eqref{eq:newconstraint}, we consider the policy class $\Pi_0 = \{\pi_\theta: \pi_\theta(1|x) = I(x>\theta), \theta \in [-1,1]\}$ which is correctly specified for all 4 scenarios, and the interpolation weight $w_t(x) = (1-t)w_0(x) + t$, where $t \in [0,1]$. To deal with the unknown $\thetatrue$, we use the first approach we suggested previously and replace $\thetatrue$ with $\hat\theta_{w_t}$ for each $t$. In addition, we divide the sample into two subsamples $D_1$ and $D_2$, and evaluate a two-fold cross-fitting procedure. In this procedure, we first use $D_1$ to form the initial estimate $\hat\theta_{w_t,1}$. We then obtain an estimate $\hat w^{\dagger}_2$ of the optimal weight by solving the sample version of \eqref{eq:newconstraint} wherein $\thetatrue$ is estimated with $\hat\theta_{w_t,1}$ and all other means and nuisance functions are estimated using data from $D_2$. The estimate of the optimal weight $\hat w^{\dagger}_1$ is obtained similarly, with the roles of $D_1$ and $D_2$ reversed. 
The final policy is then learned by maximizing $V_{n,1}(\pi;\hat w^{\dagger}_1)+V_{n,2}(\pi;\hat w^{\dagger}_2)$, where $V_{n,j}$ is an efficient estimator of the weighted value based on subsample $D_j$. For comparison, we also include the oracle where the true value of $\thetatrue$ is used in estimating the optimal weight $w^{\dagger}$ and no cross-fitting is used. For the constraint in \eqref{eq:secondconstraint}, we again use two-fold cross-fitting based on two subsamples $D_1$ and $D_2$. We obtain $\hat w^{\ddagger}_1$ by solving an estimate of \eqref{eq:secondconstraint} based on $D_2$, and obtain $\hat w^{\ddagger}_2$ by solving an estimate of \eqref{eq:secondconstraint} based on $D_1$. The final policy is learned by maximizing $V_{n,1}(\pi;\hat w^{\ddagger}_1)+V_{n,2}(\pi;\hat w^{\ddagger}_2)$.


In each simulation, we use a training sample size of 2,000 and estimate the regret with an independent test sample of size 100,000. Relevant nuisance functions are estimated by gradient boosting machines using the R package \textit{gbm} \citep{gbm} with the same settings as were used in Kallus' simulation studies. All results are based on 5,000 replications.

\begin{table}
    \centering
    \begin{tabular}{rrrrrr}
    \toprule
         & \multicolumn{5}{c}{$\hat\theta - \thetatrue$}   \\
    \cmidrule(l){2-6} 
    scenario & retarget & uniform & $w^{\dagger}$ (oracle) & $w^{\dagger}$ (CF) & $w^{\ddagger}$ (CF) \\
    \midrule
    1 & -0.237 (0.429) & -0.094 (0.265) & -0.094 (0.265) & -0.179 (0.341) & -0.172 (0.344) \\
    2 &-0.943 (0.547) & 0.000 (0.132) & 0.000 (0.132) & -0.006 (0.155) & -0.005 (0.202)\\
    3 & 0.004 (0.156) & -0.140 (0.439) &  0.004 (0.156)&  -0.141 (0.408) & -0.278 (0.549) \\
    4 & -0.077 (0.342) & 0.017 (0.331) & -0.029 (0.324) & -0.054 (0.333) & -0.039 (0.372) \\
    \toprule
    & \multicolumn{5}{c}{regret} \\
    \cmidrule(l){2-6}
    scenario & retarget & uniform & $w^{\dagger}$ (oracle) & $w^{\dagger}$ (CF) & $w^{\ddagger}$ (CF)  \\
    \midrule
    1 & 0.013 (0.020) & 0.006 (0.012) & 0.006 (0.012) & 0.009 (0.016) & 0.009 (0.017)  \\
    2 & 0.052 (0.022) & 0.004 (0.006) & 0.004 (0.006) & 0.004 (0.007) & 0.008 (0.011) \\
    3 & 0.006 (0.018) & 0.053 (0.092) & 0.006 (0.018) & 0.047 (0.090) & 0.095 (0.111) \\
    4 & 0.018 (0.030) & 0.018 (0.027) & 0.017 (0.027) & 0.017 (0.029) & 0.020 (0.032)\\
    \bottomrule
    \end{tabular}
    \caption{Performance of the retargeting weight, the uniform weight and curvature-based weights: estimation of the parameter and regret. The presented results are in the form of mean (sd). ``CF" stands for cross-fitting.}
    \label{sim}
\end{table}

Table \ref{sim} compares the performance of the retargeting, uniform, and curvature-based weights in terms of the estimation of $\theta$ and the regret. We start by discussing the performance of the retargeting and uniform weights. In scenarios 1 and 2, the retargeting weight underperforms relative to the uniform weight. The retargeting weight performs the better in scenario 3. In scenario 4, both methods have comparable performance. All of these results conform to what we would expect given the curvature in the various scenarios that we discussed in the first paragraph of this subsection. Using the curvature-based weights appears to provide a reliable means to adapt across these various scenarios. In scenarios 1, 2, and 4, these weights perform comparably to the uniform weight and, therefore, outperform the retargeting weight. In scenario 3, the cross-fitted weight with the second-derivative constraint still outperforms the uniform weight, whereas the cross-fitted weight with the global curvature constraint does not. Advantageously, the oracle weight fitted with the second-derivative constraint almost always picks $t=1$ (uniform weight) in scenarios 1 and 2, and almost always picks $t=0$ (retargeting weight) in scenario 3.

Overall, these results show that whether the policy learned using the retargeting weight outperforms that learned using the uniform weight depends on how the retargeting weight modifies the curvature of the value function at the optimal policy. 
Furthermore, they suggest that accounting for the curvature of this function when finding the weight has the potential to improve the testing performance of the learned policy.

\section{Discussion of \cite*{mo2020learning}}\label{sec:Mo}

\subsection{Alternative strategies for integrating calibration data}



As discussed in the Introduction, the new distributionally robust method provides a means of learning a policy that will perform well even at testing distributions $\mathbb{Q}$ that differ from the distribution $\mathbb{P}$ that generated the data in terms of the covariate distribution only --- to formalize this, the authors focus on testing distributions that satisfy their Assumption 1, namely that $\mathbb{Q}$ for which $\mathbb{Q}\ll \mathbb{P}$ and $d\mathbb{Q}/d\mathbb{P}$ is measurable with respect to the sigma-field generated by a randomly drawn training covariate $X$. To accomplish this, their method relies on the choice of a distributional robustness constant $c\in[1,\infty)$ and a power order $k\in(1,\infty]$. Together, these quantities determine the size and shape of the uncertainty set $\mathcal{P}_c^k(\mathbb{P}_X):= \left\{\mathbb{Q}_X\ll\mathbb{P}_X : \|d\mathbb{Q}_X/d\mathbb{P}_X\|_{L^k(\mathbb{P})}\le c\right\}$ of testing covariate distributions over which their method aims to optimize the worst-case performance. 
In their simulations, the authors found that their method was not particularly sensitive to the choice of $k$, but was quite sensitive to the choice of $c$. We, therefore, assume hereafter that $k$ is fixed and that the objective is to select the distributional robustness constant $c$.

To tune this quantity, the authors propose obtaining ``additional information to calibrate the choice of $c$ so that the [distributionally robust policy] performs well on a specific testing distribution''. In particular, they focus on the case where a random sample is available from this specific testing distribution of interest, which we denote by $\Ptest$. 
When a calibration sample from $\Ptest$ is available, the authors propose to use this data to tune the distributional robustness constant $c$ upon which their method relies. Specifically, they propose to use the available data to select $c$ so that the policy $\hat{\pi}_c$ that was learned on the training data using parameter $c$ will perform well at $\Ptest$. Performance is quantified via the value $V(\hat{\pi}_c;\wtest)$ or the centered value $R(\hat{\pi}_c;\wtest,\hat{\bar{\pi}}_c)$, where $\wtest:= d\PtestX/d\mathbb{P}_X$ and  $\hat{\bar{\pi}}_c$ is the policy that assigns each action $a\in\{-1,1\}$ with probability $\hat{\bar{\pi}}_c(a|x)=1-\hat{\pi}_c(a|x)$. The distributional robustness constant $c$ is chosen based on an estimate of this value. 
We find the idea to leverage data from a testing distribution of actual interest to be compelling. We see that using this data provides means to choose a distributional robustness constant $c$ at which their learned policy should perform about as well in the testing population as it would with the best such constant.

When we learned of this calibration procedure, we began to wonder which value of $c$ it would typically select in practice. In favorable cases, such as those where the restricted policy class contains the unconstrained optimal policy, Lemma~2.1 in Kallus suggests that the policy learned when $c=1$ may perform well even under large covariate shifts, that is, when $\|w^*\|_{L^k(\mathbb{P})}$ is large. Nevertheless, in certain less favorable cases, it seems reasonable to conjecture that the learned policy will only perform well when $c\ge \|\wtest\|_{L^k(\mathbb{P})}$, so that $\PtestX\in\mathcal{P}_c^k(\mathbb{P}_X)$. If this conjecture is correct, then it suggests that, under large covariate shifts, the distributionally robust policy estimator may be overly conservative in some cases, as the uncertainty set $\mathcal{P}_c^k(\mathbb{P}_X)$ over which a least favorable testing distribution is sought will need be large for the learned policy to perform well at the testing distribution. 

To illustrate this conjecture in the simple setting where $k=2$ and $X$ has support in $\{0,1,2\}$, we have displayed two possible configurations of $\mathbb{P}$ and $\Ptest$. On the left, there is limited covariate shift. In this case, a small value of $c$ is needed to ensure that $\PtestX\in\mathcal{P}_c^2(\mathbb{P}_X)$. Therefore, we do not expect that the proposed method will be conservative when the least favorable distribution is sought over $\mathcal{P}_c^2(\mathbb{P}_X)$ in this case. On the right, there is large covariate shift. In this case, $c$ must be chosen to be large to ensure that $\PtestX\in\mathcal{P}_c^2(\mathbb{P}_X)$. Consequently, if the minimal such $c$ is used, then a wide variety of possible $\mathbb{Q}_X$ distributions must be considered when seeking a least favorable distribution over $\mathcal{P}_c^2(\mathbb{P}_X)$. For example, $\mathcal{P}_c^2(\mathbb{P}_X)$ contains distributions $\mathbb{Q}_X$ that put most of their mass near 0, and also includes $\mathbb{Q}_X$ that put most of their mass near 2. This is undesirable in that these distributions do not resemble the actual testing covariate distribution $\PtestX$, for which $\PtestX\{X=0\}$ and $\PtestX\{X=2\}$ are both close to zero. Moreover, in the case that a small calibration sample is available from $\Ptest$, the data would clearly indicate that distributions that put most of their mass near either $0$ or $2$ need not be considered if the learned policy is only to be deployed on $\Ptest$.

\begin{figure}
    \centering
    \includegraphics[width=0.6\textwidth]{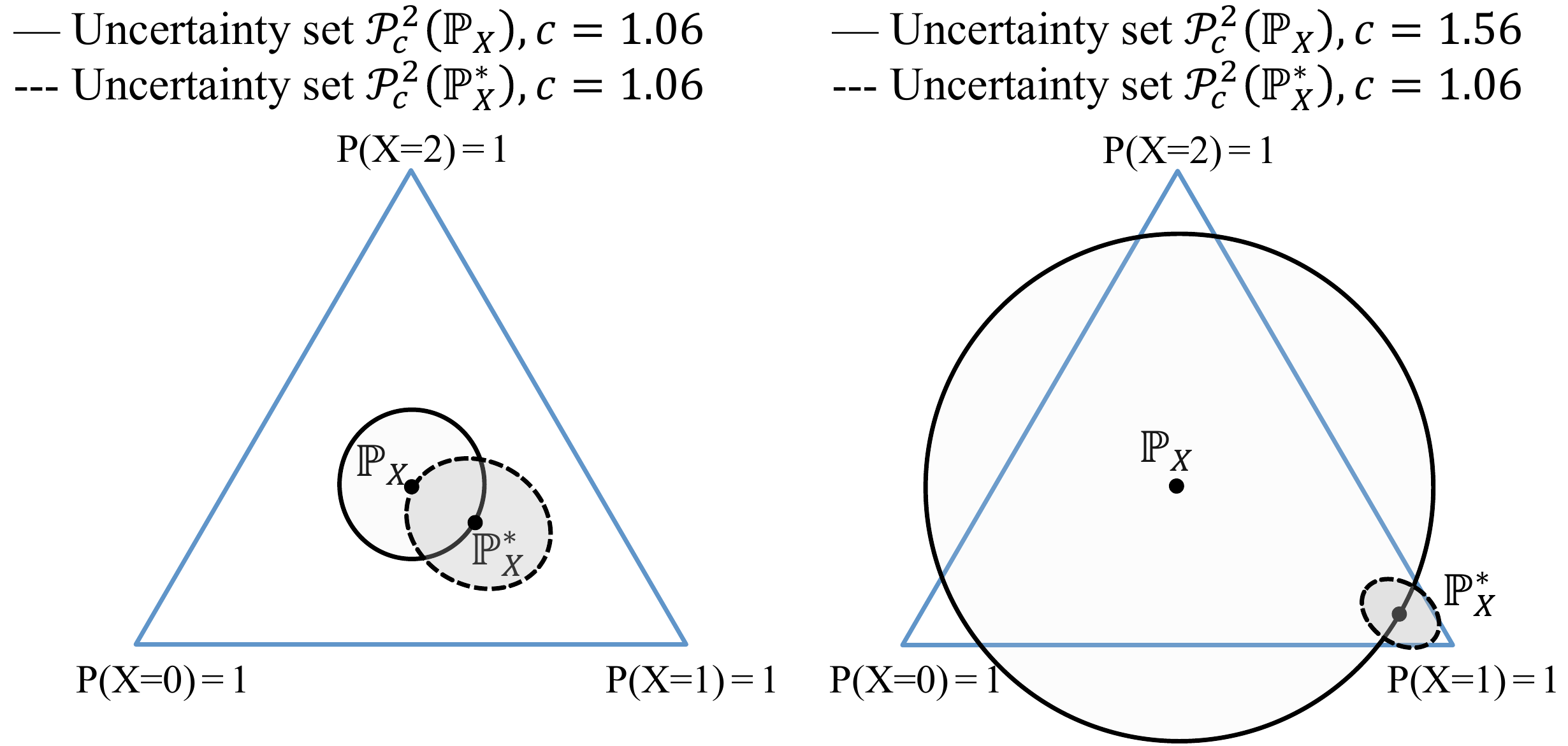}
    \caption{A comparison of uncertainty sets ($k=2$) centered at $\mathbb{P}_X=\textnormal{Unif}\{0,1,2\}$ and those centered at $\PtestX=\textnormal{Multinom}(1,p)$. Left: $p=(1/4,1/2,1/4)$. Right: $p=(1/20,9/10,1/20)$. A minimal $c$ such that $\mathcal{P}^2_c(\mathbb{P}_X)$ contains $\PtestX$ was chosen to construct $\mathcal{P}^2_c(\mathbb{P}_X)$. The uncertainty set $\mathcal{P}^2_{1.06}(\PtestX)$ is displayed for comparison.}
    \label{uncertaintyset}
\end{figure}

It is natural to wonder whether there are alternative approaches for integrating calibration data from the testing distribution $\Ptest$ of interest into a distributionally robust policy learning scheme. One possibility that we can imagine would be to center the uncertainty set at the calibration sample, so that a least favorable covariate distribution is sought over an uncertainty set $\mathcal{P}_c^k(\PtestX)$, rather than over $\mathcal{P}_c^k(\mathbb{P}_X)$. Candidate uncertainty sets of this type are shown in Figure~\ref{uncertaintyset}.  
Because $\mathcal{P}_c^k(\PtestX)$ necessarily contains the testing distribution $\PtestX$, we do not expect that it would be necessary to choose $c$ large in cases where $\Ptest$ is far from $\mathbb{P}$. How to choose $c$ in this setting is an open question. One possibility is to choose $c$ so that $\mathcal{P}_c^k(\PtestX)$ corresponds to a $1-\alpha$ confidence set for $\PtestX$ based on the calibration sample.

A similar strategy to that of the authors can be employed to learn a policy $\pi$ that performs well in terms of the $\mathcal{P}_c^k(\PtestX)$-worst-case centered value, namely $\inf_{\mathbb{Q}_X\in\mathcal{P}_c^k(\PtestX)} \E_{\mathbb{Q}_X}[\{2\pi(1|X)-1\}C(X)]$. In particular, the empirical objective presented in Section~2.6 of \citeauthor{mo2020learning} can be modified to optimize this criterion by replacing the empirical mean over the covariates $X$ observed in the training sample by an empirical mean over the covariates $X$ observed in the calibration sample. In this modified scheme, the estimate $\widehat{C}_n(\cdot)$ of $C(\cdot)$ can be obtained either using the training data or, in the case that action-outcome information is available from the calibration sample, can alternatively be obtained by pooling all available data to estimate this function. 

As we have indicated, we are unsure of the rationale behind centering the uncertainty set at the training covariate distribution $\mathbb{P}_X$ when a calibration sample from the testing distribution at which the learned policy will be deployed is available. Nevertheless, we can imagine a closely related setting where calibration data is available and we see this centering as natural. In particular, suppose that a single learned policy will be deployed at several testing distributions $\mathbb{P}_{1}^*,\ldots,\mathbb{P}_{L}^*$, and separate calibration samples are available from each of these distributions. In this case, it seems natural to select the $c$ indexing the uncertainty set $\mathcal{P}_c^k(\mathbb{P}_X)$ so that learned policy $\hat{\pi}_c$ will perform well at the least favorable of these testing distributions, that is, to select $c$ to maximize an estimate of $\min_{\ell=1,\ldots,L} V(\hat{\pi}_c;d\mathbb{P}_{\ell,X}^{*}/d\mathbb{P}_X)$. The learned policy should perform well at all $L$ known testing distributions, along with a variety of other testing distributions at which it may later be deployed.





\subsection{Efficient estimation of the testing value function}

We consider the same two calibration scenarios as the authors. In the first of these scenarios, we suppose that an iid sample of $n$ copies of $(S,X,A,Y)$ is observed, where $S$ is a Bernoulli random variable, $(X,A,Y)|S=1\sim \mathbb{P}$, and $(X,A,Y)|S=0\sim \Ptest$. We denote the distribution of $(S,X,A,Y)$ by $P$. The collections of observations with $S=1$ and $S=0$ are referred to as the training sample and calibration sample, respectively. The second scenario is the same as the first except that the action $A$ and outcome $Y$ are missing for all observations in the calibration sample. 

\begin{figure}
    \centering
    \includegraphics[scale=0.4]{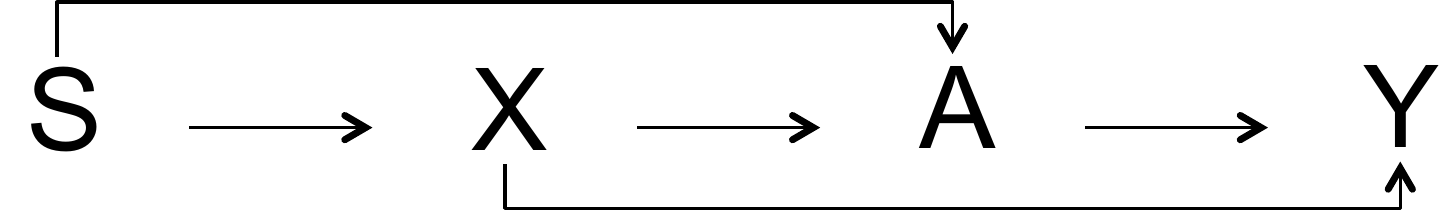}
    \caption{The directed acyclic graph illustrating Assumption 1'.}
    \label{DAG}
\end{figure}

Under this setup, Assumption 1 can be described as $(Y,A)\indep S\mid X$. As noted by the authors in their Remark 2, their method is also valid under weaker assumptions. Here we adopt one such assumption, which we refer to as Assumption~1'. In particular, we suppose that $Y\indep S\mid A,X$ --- see Figure~\ref{DAG} for the directed acyclic graph visualizing this assumption.

As described in the preceding subsection, the authors propose to make use of the calibration data to tune the distributional robustness constant $c$ via estimation of $V(\hat{\pi}_c;\wtest)$ or $R(\hat{\pi}_c;\wtest,\hat{\bar{\pi}}_c)$. The authors found in their simulations that the performance of their method improved when the precision of the value estimator improved, either due to increasing the calibration sample size or due to using a more efficient estimator.
Here we propose four alternative estimators to those proposed in \citeauthor{mo2020learning}, three for $V(\pi;\wtest)$ and one for $R(\pi;\wtest,\bar{\pi})$, where $\pi$ will be used to denote some fixed policy for the remainder of this subsection and we let $\bar{\pi}(a|x):=1-\pi(a|x)$. We use tools from semiparametric efficiency theory to argue that these estimators will generally be more efficient than those proposed in the work under discussion. These estimators may therefore provide an improved means to tune the distributional robustness constant $c$. 


To define each of these estimators, we require estimates of several nuisance functions. Estimates $\hat{\mu}(a,x)$ of the outcome regression $ \mu(a,x)$ and $ \hat{\phi}(a|x,s)$ of the propensity score $ \phi(a|x,s):=P(A=a|X=x,S=s)$ may be obtained via parametric regression or via supervised machine learning approaches. Other estimators also rely on having estimates $\widehat{P}(S=s)$ of $P(S=s)$ and $\widehat{P}(S=s|X=x)$ of $P(S=s|X=x)$. We suggest estimating $P(S=s)$ empirically and use flexible logistic regression for estimating $P(S=s|X=x)$. Moreover, noting that $\wtest(x)=P(S=1)P(S=0|x)/[P(S=0)P(S=1|x)]$, we let $\wtesthat(x):=\widehat{P}(S=1)\widehat{P}(S=0|x)/[\widehat{P}(S=0)\widehat{P}(S=1|x)]$.

We start by considering the scenario in which $(A,Y)$ is observed in the calibration sample. In this setting, the authors proposed the inverse probability weighted (IPW) estimator $\widehat{V}_{\textnormal{IPW}}^*(\pi) := \mathbb{E}_{n,0} [\pi(A|X) Y / \hat{\phi} (A|X,0)]$, where the outer expectation is the empirical expectation over observations in the calibration sample. A common way to improve this estimator is to augment it with a term that incorporates covariate information from all individuals, including those individuals with $A\not=\pi(X)$. This augmented inverse probability weighted (AIPW) estimator takes the following form:
\begin{equation}\label{eq:aipw}
    \widehat{V}_{\textnormal{AIPW}}^*(\pi) = \widehat{V}_{\textnormal{IPW}}^*(\pi) + \mathbb{E}_{n,0} \left[\sum_{a\in\mathcal{A}} \pi(a|X) \hat{\mu}(a,X)-\frac{\pi(A|X)}{\hat{\phi}(A|X,0)}\hat{\mu}(A,X)\right].
\end{equation}
The above is guaranteed to be at least as efficient as the IPW estimator if $\hat{\mu}$ is consistent and other reasonable regularity conditions are satisfied \citep[e.g.,][]{robins1994estimation}. It is also worth noting that the consistency and asymptotic normality of the above estimator for the value of $\pi$ under $\Ptest$ do not generally rely on Assumption~1' being satisfied. Indeed, in settings where this assumption is not plausible, the above still represents a reasonable estimator of the testing value provided $\hat{\mu}$ is estimated using only data from the calibration sample. In fact, under the usual $n^{-1/4}$-rate consistency conditions, the resulting AIPW estimator will be efficient among all regular and asymptotically linear (RAL) estimators in the scenario where an iid sample of $(S,X,A,Y)$ is observed and the only assumptions imposed on the distribution that generated these variates involve possible knowledge of the marginal distribution of $S$ and the conditional distribution of $A$ given $(X,S)$. That this estimator may be valid even when Assumption~1' does not hold arises, in part, because, for a fixed policy $\pi$, this estimator does not make any use of the training data.

Though the AIPW estimator's lack of dependence on the training data may make it more robust, it also suggests that this estimator may be inefficient when Assumption~1' in fact holds. Indeed, under Assumption~1', the training data can contain valuable information about the outcome regression $\mu$. It turns out to be possible to leverage tools from semiparametric efficiency theory to efficiently take advantage of this additional information. Specifically, within the model where Assumption~1' holds, the following estimator will generally be more efficient than the AIPW estimator under regularity conditions:
\begin{align}
    \widehat{V}_{\textnormal{eff}}^*(\pi) & = \E_n\left[\pi(A|X)\hat{\tau}(A,X) \left(y - \hat{\mu}(A,X)\right)\right]   + \E_{n,0}\left[\sum_{a \in \mathcal{A}}  \hat{\mu}(a,X) \pi(a|X)\right], \label{eq:Vcalibeff}
\end{align}
where $\hat{\tau}(a,x):= \widehat{P}(S=0|X=x)/\hat{\phi}(a|x,0) + \widehat{P}(S=1|X=x)\wtesthat(x)/\hat{\phi}(a|x,1)$ and $\E_n$ denotes the empirical expectation over all observed $(S,X,A,Y)$.
Specifically, under regularity conditions that include that $\hat{\mu}$, $\hat{\phi}$, and $x\mapsto \widehat{P}(S=1|X=x)$ are consistent at appropriate rates, it can be shown that this estimator has influence function
\begin{align*}
   (s,x,a,y)&\mapsto \pi(a|x)\tau(a,x) \left\{y - \mu(a,x)\right\}    + \sum_{a' \in \mathcal{A}}  \frac{1(s=0)}{P(S=0)}\mu(a',x)\pi(a'|x) - V(\pi;\wtest),
\end{align*}
where $\tau(a,x):= P(S=0|x)/\phi(a|x,0) +P(S=1|x)\wtest(x)/\phi(a|x,1) $. 
The above corresponds to canonical gradient \citep{pfanzagl1990estimation} in the model that is nonparametric up to the independence encoded in Assumption~1'. Consequently, when $\widehat{V}_{\textnormal{eff}}^*(\pi)$ has the above influence function, this estimator achieves the minimal possible variance among all RAL estimators. It is worth noting that the above remains the canonical gradient and $\widehat{V}_{\textnormal{eff}}^*(\pi)$ remains efficient if, in addition to Assumption~1', the model further encodes knowledge about either or both of the marginal distribution of $S$ and the conditional distribution of $A$ given $(X,S)$. The model that imposes Assumption~1 and is otherwise unrestricted represents one such example of a setting where $\widehat{V}_{\textnormal{eff}}^*(\pi)$ remains efficient.

We now turn to the scenario where $(A,Y)$ is missing for all observations in the calibration sample. Here the authors focus on estimating the centered value $R(\pi;\wtest,\bar{\pi})$. It is worth noting that, because the directed acyclic graph in Figure~\ref{DAG} does not imply any testable assumptions on the statistical model where $(A,Y)$ is missing whenever $S=0$, making this assumption will not lead to any gain in efficiency in this setting. Nevertheless, this assumption plays a critical role --- in particular, under this assumption, for any policy $\rho$ it is possible to identify $V(\rho;\wtest)$ with a parameter of the conditional distribution of $X$ given $S=0$ and of the conditional distribution of $(A,Y)$ given $S=1$ and $X$. Indeed, $V(\rho;\wtest)=\E\left[\sum_{a\in\mathcal{A}}\rho(a|X)\E[Y|A=a,X,S=1] \middle|S=0\right]$. As $R(\pi;\wtest,\bar{\pi})=V(\pi;\wtest)-V(\bar{\pi};\wtest)$, this centered value can be similarly identified. 
\citeauthor{mo2020learning} suggested a plug-in estimator, namely $\widehat{R}_{\textnormal{plugin}}^*(\pi):=\E_{n,0}[\{2\pi(1|X)-1\}\widehat{C}_n(X)]$ where $\widehat{C}_n$ is an estimate of $C$ based on the training data. This estimator has the advantage that it is easy to explain and implement, but it has the disadvantage of generally failing to be RAL, and therefore potentially having large bias compared to some alternative estimators. Using tools from semiparametric efficiency theory, we derived the alternative estimator $\widehat{R}_{\textnormal{onlyX}}^*(\pi):=\widehat{V}_{\textnormal{onlyX}}^*(\pi)-\widehat{V}_{\textnormal{onlyX}}^*(\bar{\pi})$, where, for any policy $\rho$ and letting $\E_{n,1}$ denote the empirical expectation over the training sample, we define
\begin{align*}
\widehat{V}_{\textnormal{onlyX}}^*(\rho)&:= \E_{n,1}\left[\frac{\rho(A|X) }{\hat{\phi}(A|X,0)} \wtesthat(X)\left\{Y - \hat{\mu}(A,X)\right\}\right] + \E_{n,0}\left[\sum_{a \in \mathcal{A}}  \hat{\mu}(a,X) \rho(a|X)\right].
\end{align*}
Under regularity conditions that include the consistency of $\hat{\phi}$, $\hat{\mu}$, and $\wtesthat$ 
at appropriate rates, the above estimator will be asymptotically linear with influence function
\begin{align*}
    (s,x,a,y) &\mapsto  \frac{\rho(a|x) }{\phi(a|x,0)}h_1(s) \wtest(x)\left\{y - \mu(a,x)\right\} +  \sum_{a' \in \mathcal{A}}  h_0(s)\mu(a',x) \rho(a'|x) - V(\rho;\wtest),
\end{align*}
where $h_u(s):=1(s=u)/P(S=u)$. Within the nonparametric model in which $(A,Y)$ is missing for all observations in the calibration sample, the above is the canonical gradient of the functional that returns $V(\rho;\wtest)$. Consequently, under regularity conditions, $\widehat{V}_{\textnormal{onlyX}}^*(\rho)$ is efficient among all RAL estimators of $V(\rho;\wtest)$ in this model. Similarly, under conditions, $\widehat{R}_{\textnormal{onlyX}}^*(\pi)$ is an efficient estimator of $R(\pi;\wtest,\bar{\pi})$.


Interestingly, $\widehat{R}_{\textnormal{onlyX}}^*(\pi)$ is also a valid estimator of $R(\pi;\wtest)$ in the first scenario that \citeauthor{mo2020learning} considered, namely that where $(A,Y)$ is observed on all individuals. Nevertheless, the estimator $\widehat{R}_{\textnormal{eff}}^*(\pi):=\widehat{V}_{\textnormal{eff}}^*(\pi)-\widehat{V}_{\textnormal{eff}}^*(\bar{\pi})$ will generally be more efficient in this scenario. This observation aligns with our intuition that the more outcome data collected, the better will be our estimate of the outcome regression $\mu$. It is also worth noting that, by comparing the asymptotic variances of the two estimators, it can be shown that the relative efficiency of $\widehat{R}_{\textnormal{eff}}^*(\pi)$ compared to $\widehat{R}_{\textnormal{onlyX}}^*(\pi)$ improves as the size of the calibration sample grows.

We conclude by noting that, in practice, for either of the scenarios considered above, we suggest always trying to obtain a larger calibration sample for better estimation and using cross-fitting to avoid over-fitting. When training data is incorporated into the procedure and the goal is to evaluate the value of $\hat{\pi}_c$, we recommend cross-fitting the policy $\hat{\pi}_c$ as well, so that this policy is not estimated on the same training data as is used to evaluate the empirical means that appear in the definitions of our proposed estimators. 

\subsection{Simulation study}

We now present simulation results to compare the performances of the proposed estimators. We follow the simulation setup described by Mo et al. For the training data generating process, $n=1\textnormal{,}000$, $p=10$, $X \sim N(0_p, I_p)$, $A|X \sim$ Uniform[-1,1] and $Y|A,X \sim N[m_0(X) + 0.5A\, C(X),1]$, 
where $m_0(x) := 1+ \frac{1}{p} \sum_{i=1}^p x_i$ and $C(x) = x_2 - (x_1^3 -2x_1)$. We consider the testing distribution $X \sim N(\gamma, I_p)$, where $\gamma_j=0$ for all $j=3,\ldots,p$ and the values of $(\gamma_1,\gamma_2)$ determine whether there is any covariate shift. We consider the cases of no covariate shift, so that $(\gamma_1,\gamma_2)=(0,0)$, and covariate shift, specifically with $(\gamma_1,\gamma_2) =(0.734,1.469)$. For each scenario, we compare the performances of estimators of the testing value of $\pi$ across different calibration sample sizes $n_{\textnormal{calib}}$ ranging from 50 to 1,000. 

We estimate the testing value $V(\pi;\wtest)$ of the policy $\pi(x) = 1\{x_2 - (x_1^3 -2x_1) > 0\}$. In particular, we consider the following estimators: (1) the IPW estimator proposed by \citeauthor{mo2020learning}, namely $\widehat{V}_{\textnormal{IPW}}^*(\pi)$, (2) the AIPW estimator, namely $\widehat{V}_{\textnormal{AIPW}}^*(\pi)$, (3) the estimator that is efficient under Assumption 1', namely $\widehat{V}_{\textnormal{eff}}^*(\pi)$, and (4) the estimator that is efficient among all RAL estimators that make no use of the action-outcome information in the calibration sample, namely $\widehat{V}_{\textnormal{onlyX}}^*(\pi)$. In addition, we also consider the case where all actions and outcomes are missing in the calibration sample and the goal is to estimate the centered value function $R(\pi;\wtest,\bar{\pi})$. For this scenario, the following estimators are considered: (5) the plug-in estimator proposed by \citeauthor{mo2020learning}, namely $\widehat{R}_{\textnormal{plugin}}^*(\pi)$, using the causal forest \citep{wager2018estimation} estimator for $\widehat{C}_n$ that they used in their simulations; and (6) the efficient estimator in this missing data setting, namely $\widehat{R}_{\textnormal{onlyX}}^*(\pi)$. In our upcoming table we call (3) ``eff'', (4) and (6) ``onlyX'', and (5) ``grf''.

We fit a logistic regression model to estimate $P(S=1|X=\mycdot)$ and use sieve estimators with a polynomial basis to estimate the outcome regression, where the number of terms is selected via 5-fold cross-validation. The true propensity is used for $\hat{\phi}$, which is reflective of what can be done in a randomized trial setting. We estimate $C(\cdot)$ via the causal forest \citep{wager2018estimation} implemented in the R package \textit{grf} \citep{athey2019generalized}, using 5,000 trees and selecting all tunable parameters via the package's default cross-validation settings.

The mean squared errors (MSEs) of these estimators are shown in Table~\ref{tab:MoSims}. 
The squared bias only contributes negligibly ($<1\%$) to the MSE for all estimators except for the causal forest, for which the squared bias contributes between 80\% to 95\% of the MSE. We start by considering the results under no covariate shift. In this setting, $\widehat{V}_{\textnormal{eff}}^*(\pi)$ outperforms all of the others, as anticipated by theory. As is evident from its definition in \eqref{eq:Vcalibeff}, this estimator cannot be evaluated when $(A,Y)$ are missing in the calibration sample. Interestingly, the estimator $\widehat{V}_{\textnormal{onlyX}}^*(\pi)$ that makes no use of the action-outcome information in the calibration sample is only slightly outperformed by $\widehat{V}_{\textnormal{eff}}^*(\pi)$, and dramatically outperforms both the IPW and AIPW estimators.
Overall the improvements of these two estimators over the IPW and AIPW estimators are substantial, especially when the calibration sample is small. This occurs because both the IPW and AIPW estimators only make use of $(X,A,Y)$ in the calibration sample, whereas the others also make use of data in the training sample. 

\begin{table}
\centering
\begin{adjustbox}{width=\textwidth}
\begin{tabular}{lrrrrrrrrrrrr}
\toprule
 & \multicolumn{6}{c}{no covariate shift} 
& \multicolumn{6}{c}{covariate shift}\\
\cmidrule(l){2-7} \cmidrule(l){8-13}
 & \multicolumn{4}{c}{$V(\pi;\wtest)$} 
& \multicolumn{2}{c}{$R(\pi;\wtest,\bar{\pi})$}& \multicolumn{4}{c}{$V(\pi;\wtest)$} 
& \multicolumn{2}{c}{$R(\pi;\wtest,\bar{\pi})$}\\
\cmidrule(l){2-5} \cmidrule(l){6-7}\cmidrule(l){8-11} \cmidrule(l){12-13}
$n_{\textnormal{calib}}$ &  IPW & AIPW & onlyX &  eff & grf & onlyX & IPW & AIPW & onlyX &  eff & grf & onlyX \\
\midrule
50 &0.162 & 0.154 & 0.035 & 0.035 & 0.262 & 0.126 & 0.337 & 0.180 & 0.117 & 0.095 & 1.238 & 0.374\\
100 &0.079 & 0.042 & 0.017 & 0.017 & 0.238 & 0.060 & 0.177 & 0.070 & 0.108 & 0.051 & 1.188 & 0.248\\
200 &0.041 & 0.018 & 0.010 & 0.009 & 0.230 & 0.033 & 0.087 & 0.033 & 0.050 & 0.027 & 1.181 & 0.135\\
500 & 0.017 & 0.008 & 0.005 & 0.005 & 0.225 & 0.015 & 0.036 & 0.013 & 0.043 & 0.012 & 1.147 & 0.092\\
1000 &0.008 & 0.003 & 0.004 & 0.002 & 0.221 & 0.010 & 0.018 & 0.007 & 0.034 & 0.006 & 1.136 & 0.079\\
\bottomrule
\end{tabular}
\end{adjustbox}
\caption{\label{tab:MoSims}  Mean squared error across 1,000 replications. }
\end{table}

We now turn to the case of covariate shift. Because there is only so much that can be learned from the training sample in this setting, the AIPW estimator performs about as well as  $\widehat{V}_{\textnormal{eff}}^*(\pi)$ when the calibration sample size is sufficient. At a training to calibration sample size ratio of 20 to 1, the efficient estimators can achieve MSE that is less than 30\% of that of the IPW estimator. For estimating the centered value function, using the causal forest plug-in estimator trades off larger bias for smaller variances, and results in significantly higher MSE than does using the RAL estimator $\widehat{R}_{\textnormal{onlyX}}^*(\pi)$.

{\bibliographystyle{abbrvnat}\singlespacing\small
\nocite{kallus2020more} 
\nocite{mo2020learning} 
\bibliography{references}
}

\end{document}